\documentclass[10pt,twocolumn,letterpaper]{article}

\usepackage{cvpr}
\usepackage{times}
\usepackage{epsfig}
\usepackage{graphicx}
\usepackage{amsmath}
\usepackage{amssymb}
\usepackage{tabularx}
\usepackage{multirow}
\usepackage[normalem]{ulem}
\usepackage[font=small,skip=-4pt,labelsep=period]{caption}
\usepackage[pagebackref=true,breaklinks=true,letterpaper=true,colorlinks,bookmarks=false]{hyperref}

\cvprfinalcopy % *** Uncomment this line for the final submission

\def\cvprPaperID{490} % *** Enter the CVPR Paper ID here

\ifcvprfinal\fi
\begin{document}

	%%%%%%%%% TITLE
	\title{ViP-CNN: Visual Phrase Guided Convolutional Neural Network}

	\author{Yikang Li$^{1}$, Wanli Ouyang$^{1,2}$, Xiaogang Wang$^{1}$,  Xiao'ou Tang$^{3,1}$\\
		$^{1}$The Chinese University of Hong Kong, Hong Kong SAR, China\\ 
		$^{2}$University of Sydney,   Australia \ \ \ \ \  $^{3}$Shenzhen Institutes of Advanced Technology, China \\
		{\tt\small ykli, wlouyang, xgwang@ee.cuhk.edu.hk}\ \ \ {\tt\small xtang@ie.cuhk.edu.hk} \\
	}

	\maketitle
	%\thispagestyle{empty}

	%%%%%%%%% ABSTRACT
	 \begin{abstract}
		As the intermediate level task connecting image captioning and object detection, visual relationship detection started to catch researchers' attention because of its descriptive power and clear structure. It detects the objects and captures their pair-wise interactions with a subject-predicate-object triplet, \eg $\langle person$-$ride$-$horse\rangle$. In this paper, each visual relationship is considered as a phrase with three components. We formulate the visual relationship detection as three inter-connected recognition problems and propose a Visual Phrase guided Convolutional Neural Network (ViP-CNN) to address them simultaneously. In ViP-CNN, we present a Phrase-guided Message Passing Structure (PMPS) to establish the connection among relationship components and help the model consider the three problems jointly. Corresponding non-maximum suppression method and model training strategy are also proposed. Experimental results show that our ViP-CNN outperforms the state-of-art method both in speed and accuracy. We further pretrain ViP-CNN on our cleansed Visual Genome Relationship dataset, which is found to perform better than the pretraining on the ImageNet for this task.
	\end{abstract}

	%%%%%%%%% BODY TEXT
	
	\section{Introduction}

	% Why we focus on visual phrase detection
	Booted by the development of Deep Learning, letting the computer understand an image seems to be increasingly closer. With the research on object detection gradually becoming mature~\cite{faster_rcnn, YOLO, SSD, kang2016object, kang2017object, kang2016t}, increasingly more researchers put their attention on higher-level understanding of the scene~\cite{densecap, visual7w, vqa, yang2015stacked, you2016image, show_attend_tell, lstm_cv, structured_feature_pose, chu2016crf, yang2016dense}. As an intermediate level task connecting the image caption and object detection, visual relationship/phrase detection is gaining more attention in scene understanding~\cite{visual_relationship, visual_phrase, visual_phrase_new}. 
	
\begin{figure}[t]
		\begin{center}
			\includegraphics[width=0.9\linewidth]{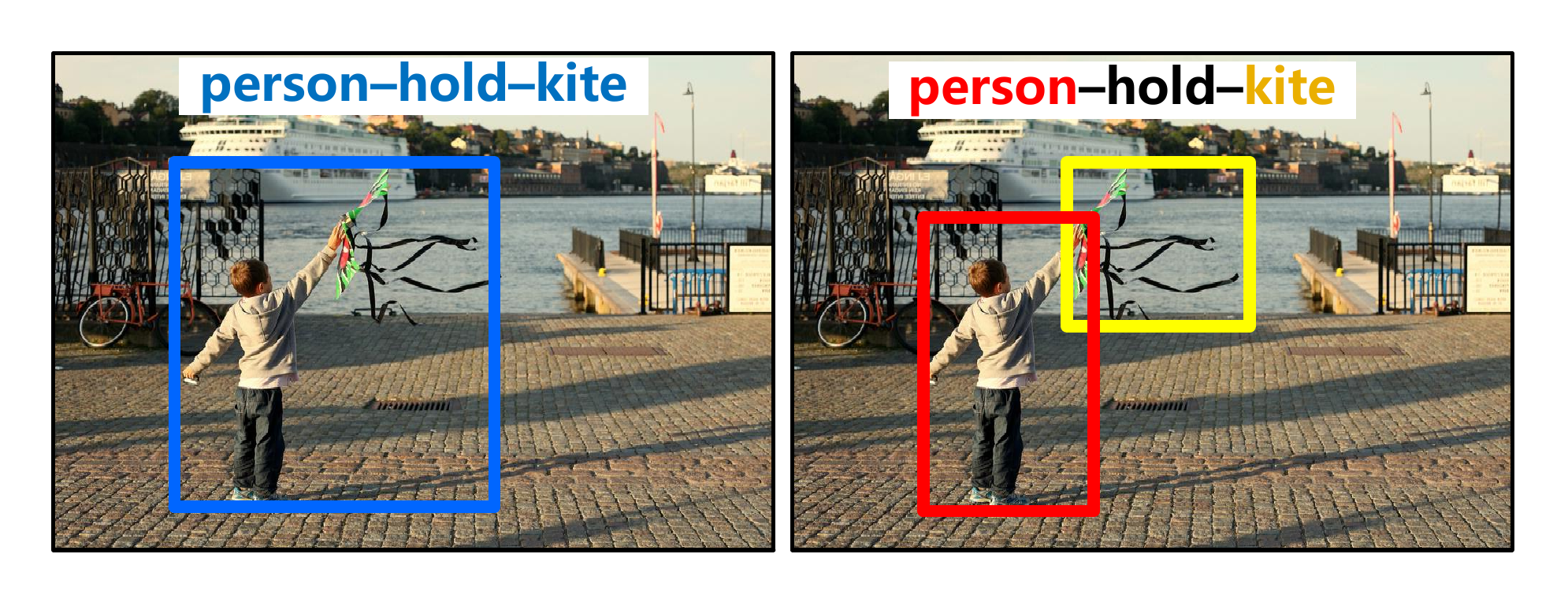}
		\end{center}
		\caption{Visual phrase (left) and visual relationship (right).}
		\label{fig:examples}
		\vspace{-12pt}
	\end{figure}

	% what is visual phrase
	\emph{Visual phrase} detection is the task of localizing a $\langle subject$-$predicate$-$object \rangle$ phrase, where predicate describes the relationship between the subject and object.
	\emph{Visual relationship} detection involves detecting and localizing pairs of interacting objects in an image and also classifying the predicate or interaction between them. For the example of $\langle person$-$hold$-$kite \rangle$ in Figure~\ref{fig:examples}, visual relationship detection aims at locating the person, the kite and classifying the pair-wise relation `hold', while visual phrase concentrates on describing the region as an integrated whole.  
	
	Compared with localized objects, visual relationships have more expressive power. It can help to build up the connection between objects within the image, which opens a new dimension for image understanding.
	In addition, unlike captions, the fixed structure of visual relationships makes it possible to explicitly design the architecture of the model for better use of the domain knowledge. Therefore, we can get a richer semantic understanding of the image through detecting visual relationships.
	
	For visual relationship detection, there are two commonly used pipelines, the bottom-up and the top-down. The bottom-up design first detects  objects and then recognizes the possible interactions among them, which is adopted by the state-of-art method~\cite{visual_relationship}. The top-down design detects the $\langle subject$-$predicate$-$object \rangle$ phrase simultaneously by regarding the relationship as an integrated whole. In comparison, the sequential order of the bottom-up implementation cuts off the feature-level connection between the two steps, which is important for correctly recognizing the relationship. Thus, we follow the top-down design by viewing the visual relationship as a phrase and solve it as three closely-connected recognition problems. Because of the joint training for the subject, predicate and object, our model is capable of learning specific visual patterns for the interaction and taking the visual interdependency into consideration. 
	
	Another motivation in our model is that the predictions of different phrase components are dependent on each other at the visual feature level. %These dependencies appear in two aspects, language level interdependencies and visual interdependencies. The former one reflects the cooccurrence of phrase components, which helps to rule out the combinations violating the common sense, like $\langle horse$-$feed$-$person \rangle$.
    This corresponds to learning special features for the pair-wise relationships. For example, visual connection of the subject (person) appearing sitting on something and an object (sofa) with the appearance of human's legs on it help to enhance the evidence of the predicate ``sit on''. In return, the specific visual features for ``sit on'' also help to inference the subject~(person) and object~(sofa) as well. They collectively help to detect the relationship, $\langle person$-$sit\ on$-$sofa \rangle$. Therefore, how to use utilize such dependencies would be the core problem of visual relationship detections.

	Based on the analysis above, we propose Phrase-guided Message Passing Strucvture (PMPS) to model the connection among relationship components.  It can be implemented with convolutional layers or fully-connected layers. Deep networks typically pass information feedforward. In contrast, PMPS extracts useful information from other components in the phrase to refine the features before going to the next level. With the sense of other components introduced by PMPS, our model can predict the subject, object and predicate simultaneously as an integrated whole. Besides, unlike simply widening the network, PMPS makes good use of the domain knowledge of the problem and the message passing strategies are designed according to the specific structure of phrases. Such explicit design reduces the parameter number and makes the training easier. 
	%With PMPS, deep models for s/o/p can improve one another by extracting useful information from other models to refine features before going to the next layer. It is faster and more accurate than previous methods. 

	% contribution
	Our main contributions of our work are summarized as three-fold:
	
	First, we propose a phrase-guided visual relationship detection framework, which can detect the relationship in one step. Corresponding non-maximum suppression method and training strategy are also proposed to improve the speed and accuracy of our model.
	
	Second, we propose Phrase-guided Message Passing Structure (PMPS) and corresponding training strategy to leverage the interdependency of three models for more accurage recognition.  In PMPS, a new gather-broadcast message passing flow mechanism is proposed, which can be applied to various layers across deep models.
	
	Third, we investigate two ways of utilizing Visual Genome Relationship dataset~\cite{visual_genome} to pretrain ViP-CNN, both of which improve the performance of our model when compared with the pretraining on ImageNet. 
	
	On the benchmark dataset~\cite{visual_relationship}, our approach is, respectively, 13.48\% and 6.88\% higher for visual phrase detection and  visual relationship detection task compared with the state-of-art method.
	
	\begin{figure*}[t]
	\vspace{-5pt}
	\begin{center}
		\includegraphics[width=0.9\linewidth]{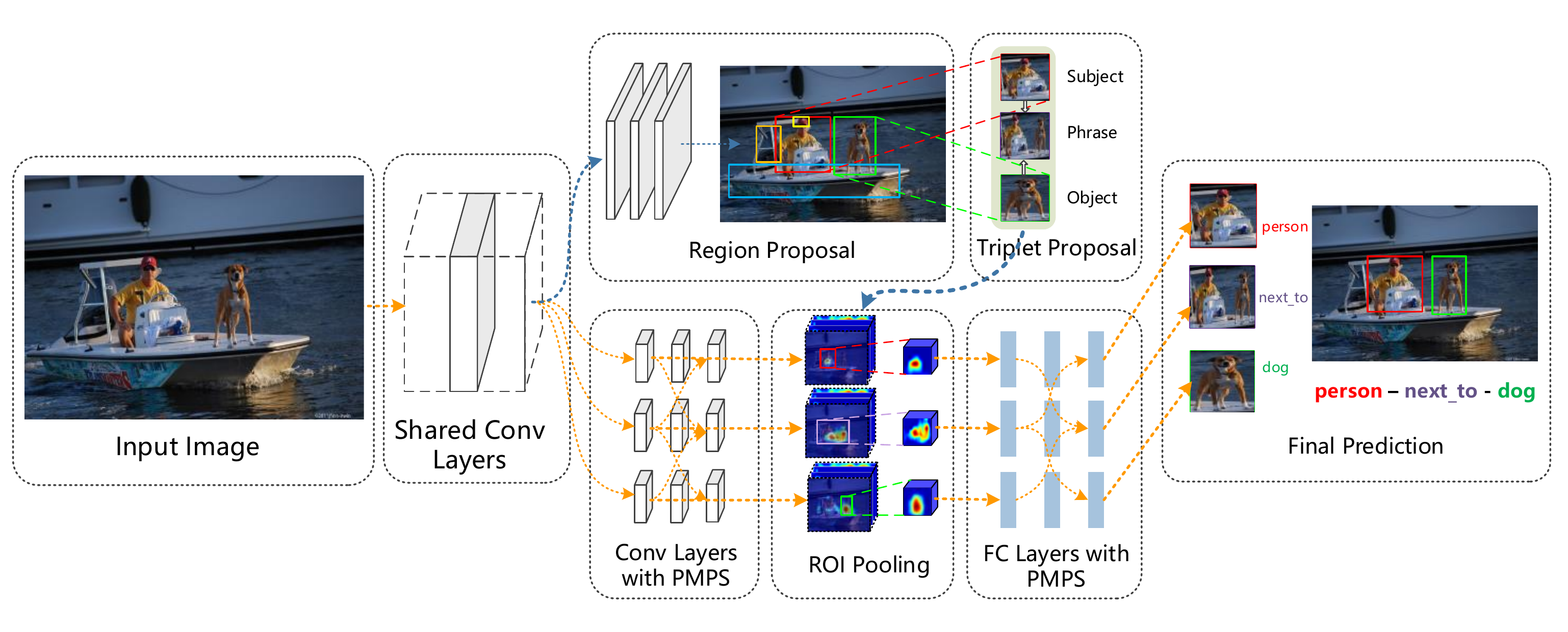}
	\end{center}
	\caption{Overview of ViP-CNN. It generates triplet proposals with RPN and then feeds them into corresponding models. The three branches are interconnected at both  conv layers and fc layers using our proposed Phrase-guided Message Passing Structure (PMPS). Results from three branches make the final prediction simultaneously. }
	\label{fig:pipeline}
	\vspace{-10pt}
\end{figure*}

	\section{Related Work}

	As the intermediate level task connecting image captioning and object detection, visual relationship detection is rooted in object detection, but shares many properties with image captioning. Also, our proposed model also involves the message passing structures. We review related works on these topics.
	
	\textbf{Object Detection:} As the foundation of image understanding, object detection has been investigated for years. Convolutional Neural Networks~\cite{imagenet_hinton} were first introduced by the R-CNN~\cite{rcnn} for object detection. It processes the regions of interest independently, which is time-consuming. Then SPP-net~\cite{spp_net}, Fast R-CNN ~\cite{fast_rcnn} were proposed to share convolutional layers among regions in classification. Ren, \etal  proposed Faster R-CNN by utilizing CNN to do region proposal~\cite{faster_rcnn}. YOLO~\cite{YOLO} and SSD~\cite{SSD} shared more convolutional layers for region proposal and region classification and made detection even faster. ViP-CNN is based on Faster R-CNN due to its superior performance. However, it is not trivial to arrange the subject, predicate and object into an end-to-end framework.  We propose PMPS and corresponding training scheme that uses the entire phrase to guide learning. The entire design is based on careful analysis of the specific problem.
	
	\textbf{Image Caption:} Describing image with natural language have been explored for many years~\cite{barnard2003matching, farhadi2010every, jia2011learning, kulkarni2013babytalk, kuznetsova2013generalizing, socher2010connecting}. Recently, using the visual features from CNN, Recurrent Neural Networks (RNNs)~\cite{rnn, lstm} have been adopted to generate captions because of its success on processing natural language. Combining the RNN and CNN becomes a standard pipeline on solving the Image Captioning problems~\cite{show_attend_tell, densecap, chen2014learning, donahue2015long, fang2015captions, karpathy2015deep}. However, the pipeline does not fit for visual relationship detection due to the difference between the sentence for image captioning and phrase for visual relationship detection. Compared to sentences, the phrase has fixed structure of subject-predicate-object. In addition, most of the related works focus on the whole image or image region, while the visual relationship detection targets on the region and its subregions.

	\textbf{Visual Relationship Detection:} Visual relationships are not a new concept. Sadeghi, \etal has proved the phrase, as a whole, can facilitate object recognition because of its special visual appearance~\cite{visual_phrase}. Desai, \etal used the phrase that describes the interaction between a person and objects to facilitate actions, pose and object detection~\cite{desai2012detecting}. However, most of the existing works were done for leveraging the relationship for other tasks~\cite{gupta2008beyond, kumar2010efficiently, russell2006using}. Rohrbach, \etal also investigated grounding the free-form phrases by reconstruction a given phrase using the attention mechanism~\cite{rohrbach2015grounding}. Lu, \etal first formalized the visual relationship detection as a task and proposed the state-of-art method by leveraging the language prior to model the correlation between subject/object and predicate~\cite{visual_phrase}.  However, these methods do not consider the visual feature level connection among the subject, predicate and object, which is used in our approach to improve the detection accuracy.

	\textbf{Message Passing Structure:} Deformable parts models are used for modeling the relationships among objects and their parts~\cite{dpm_cnn, deepid_net, desai2012detecting}. Conditional Random Field (CRF) is also proved a powerful tool for dependency learning~\cite{CRF_as_rnn, koltun2011efficient}. However, these works mainly focus on the correlations among predicted labels or object parts. Recurrent Neural Network (RNN) passes the information at feature level ~\cite{rnn, lstm} in sequential order. The fixed flow path and sharing of model parameters in RNN do not fit for the structural information in the phrase detection. Therefore, we propose a phrase-guided message passing structure with specific message passing flow to model the relationship among the visual features in the subject, predicate and object.

	\section{ViP-CNN}
	\subsection{Overview}
	An overview of our proposed model is shown in Figure \ref{fig:pipeline}. VGG-Net~\cite{simonyan2014very} is used as the basic building block for our ViP-CNN. Our model divides the entire procedure into two parts: triplet proposal and phrase recognition. Inspired by Faster R-CNN~\cite{faster_rcnn}, triplet proposal and phrase recognition can share most of the convolutional layers to make the inference much faster.
	
	ViP-CNN takes an entire image as input, whose shorter side is scaled to 400 with aspect ratio kept. The image is fed into several convolutional (conv) and max-pooling layers to produce the feature map, which corresponds to Conv1\_1 to Conv4\_3 in the VGG-Net. Then the network is split into four branches. One for triplet proposal and three for phrase detection.
	
	\emph{Triplet proposal branch.}
	Taking the output of the Conv4\_3 as input, three convolutional layers are used for extracting CNN features.
	Then features are used for proposing class-free regions of interest (ROIs) using the approach of RPN~\cite{faster_rcnn}.
	By grouping these ROIs, triplet proposal is obtained. A triplet, denoted by $(\mathbf{b}_s, \mathbf{b}_p, \mathbf{b}_o)$, is made up of three ROIs, subject ROI ($\mathbf{b}_s$), object ROI ($\mathbf{b}_p$) and predicate ROI ($\mathbf{b}_o$). The predicate ROI is the box that tightly covers both the subject and the object. Due to the sparsity of relationship annotations, triplet non-maximum suppression (triplet NMS) is proposed to reduce the redundancy. The remaining triplets are used for the phrase recognition branch. For the example in Figure \ref{fig:pipeline}, the triplet contains the bounding box for the subject (person), the object (dog), and their union.
	
	\emph{Phrase detection branches.}
	Taking the output of the Conv4\_3 as input, subject, predicate, and object have their own models because they have different visual appearances and different statuses in the relationship. %These branches are called subject branch, predicate branch and object branch. 
	We use the subject branch for illustration, which applies for the other two.
	Three convolutional layers for the subject branch are used on the shared feature map. Then the subject proposal is fed into ROI pooling layer~\cite{fast_rcnn} to obtain corresponding features with fixed size. Features are processed by several subsequent fully-connected (fc) layers for category estimation and bounding box regression. The confidence of $\langle subject$-$predicate$-$object \rangle$ phrases is measured by the product of scores for subject, object and predicate. ViP-CNN is capable of detecting both visual phrases and visual relationships by adopting different bounding box regression values. 
	Moreover, the phrase guided message passing structure is introduced to exchange the information among models, which helps to leverage the connection among the three components for better results.

	\subsection{Triplet Proposal with NMS}
	
	First, we use RPN~\cite{faster_rcnn} to generate object proposals based on the feature map. RPN is trained using both subject and object bounding boxes. Proposals are sorted in the order of objectiveness scores and top $ N $ proposals are selected. We group the proposals to construct $ N^{2} $ $ \langle subject \text{ - } object \rangle $ pairs. The predicate ROI is the union of the subject ROI and the object ROI. Thus we have $ N^{2} $ $ \langle subject \text{-} predicate \text{-} object \rangle$ triplet proposals. 

	\begin{figure}[t]
		\begin{center}
			\includegraphics[width=.8\linewidth]{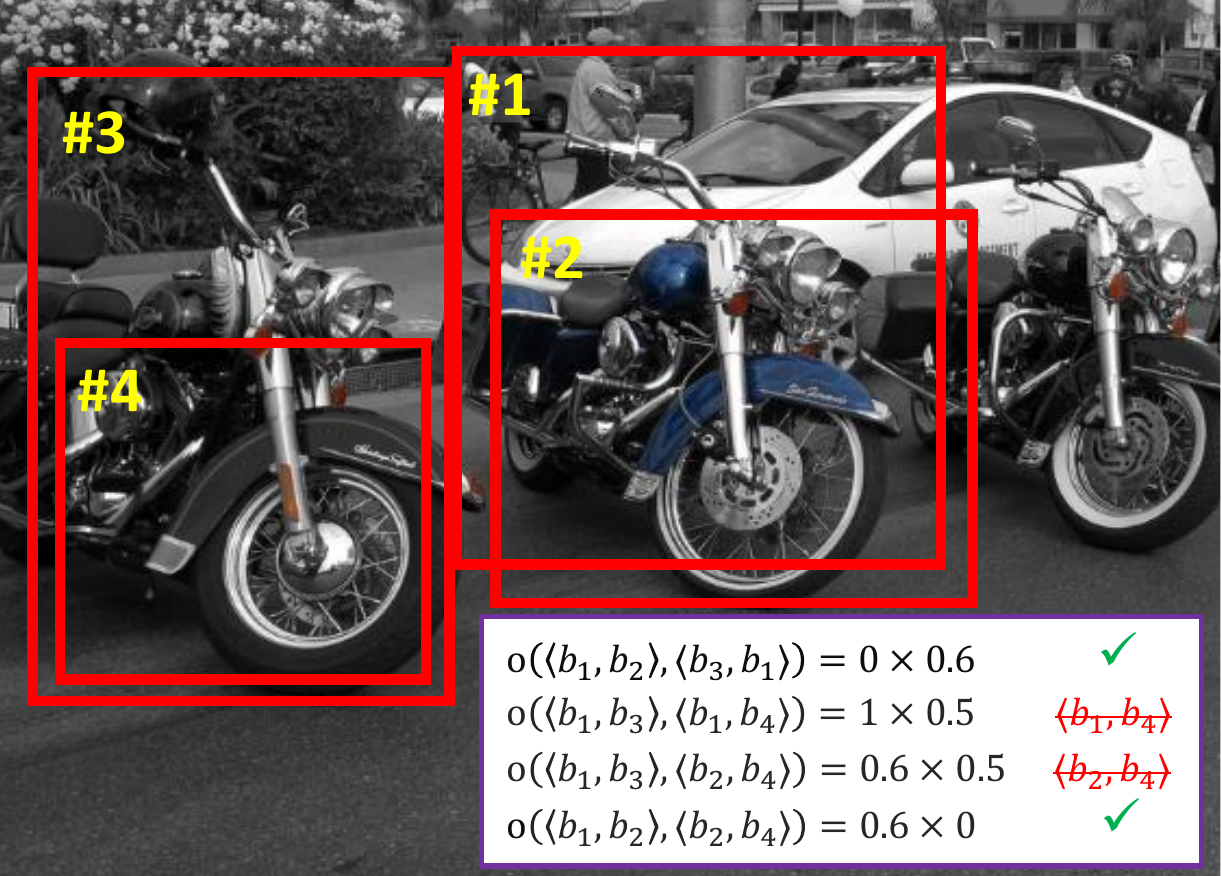}
		\end{center}
		\caption{Illustration of our triplet NMS. Smaller index means higher objectiveness score. Some pairs are  selected for illustration. $ \mathbf{b}_i $ means the \#i ROI, and $\langle\mathbf{b}_1, \mathbf{b}_2 \rangle$ represents the $\langle$subject, object$\rangle$ ROI pairs. Assume $ o\left(\mathbf{b}_1, \mathbf{b}_2\right) =0.6 $, $ o\left( \mathbf{b}_3, \mathbf{b}_4\right) =0.5 $ and the triplet NMS threshold is 0.25.  }
		\label{fig:nms}
		\vspace{-5pt}
	\end{figure}

	Under this settings, 300 object proposals will generate 90,000 triplet proposals, which is a heavy load for the subsequent relationship detection step. It also deteriorates the final result measured by Recall because of the redundancy in the triplet proposals with high overlap. Therefore, we propose a triplet non-maximum suppression (NMS) method to reduce the redundant triplet proposals (figure \ref{fig:nms}). Triplet NMS is done previous to the detection procedure for the sake of inference speed.
	
	Similar to the NMS in object proposal, triplet NMS is also based on the overlap and the objectiveness score, which is introduced below.
	
	Denote the ROI triplet by $ \mathbf{t}_i=\langle \mathbf{b}_{s, i}, \mathbf{b}_{p, i}, \mathbf{b}_{o, i} \rangle$, where $\mathbf{b}_{s, i}, \mathbf{b}_{p, i}$ and $\mathbf{b}_{o, i}$ are, respectively, the bounding boxes for subject, predicate and object.  Denote $ o\left(\mathbf{b}_{*,1}, \mathbf{b}_{*,2} \right)$ as the area of intersection between $ \mathbf{b}_{*,1} $ and $ \mathbf{b}_{*,2} $ divided by the area of their union. Then, the triplet overlap $o(\mathbf{t}_{1}, \mathbf{t}_{2})$ is the product of $ o(\mathbf{b}_{s, 1}, \mathbf{b}_{s, 2}) $ and $ o(\mathbf{b}_{o, 1}, \mathbf{b}_{o, 2}) $.
	The objectiveness score of the triplet is the product of subject objectiveness score and object objectiveness score. 
	
	With the triplet overlap and objectiveness score defined, greedy NMS~\cite{rcnn} is done to remove redundant triplet proposals. In the experiment, we use  250 object ROIs to produce triplets and set 0.25 as the triplet NMS threshold. Under this setting, the number of triplets will be reduced from 62,500 to about 1,600, which increases the speed by more than 20 times.
	
	\subsection{Phrase-guided Message Passing Structure}

	For normal CNN, only visual intra-dependencies are considered, where features at level $l$ of the subject, predicate and object as $\mathbf{h}_{s}^l$,  $\mathbf{h}_{p}^l$, and $\mathbf{h}_{o}^l$ respectively, are obtained as below:
	\begin{equation}
	\begin{split}
	\mathbf{h}_*^l = f\left(\mathbf{W}_*^l \otimes \mathbf{h}_*^{l-1} +  \mathbf{b}_*^{l}\right)		
	\end{split}
	\label{eq:CNNbranch}
	\end{equation}
	where $*$ can be $\left\{s, p, o\right\}$ and $ \otimes $ denotes the matrix-vector product for fc layers and convolution for conv layers. $\mathbf{W}_*^l$ and $\mathbf{b}_*^{l}$ are parameters of fc or conv layers. 
	
	Under this settings, connections among subject, predicate and object are omitted. To employ the complementary information provided by other models, Phrase-guided Message Passing Structure (PMPS)  is proposed
	PMPS extracts useful information from the source branch to refine features of the destination branch. Passing message  horizontally across deep models helps them improve one another.
	\begin{figure}[t]
		\begin{center}
			\includegraphics[width=1.\linewidth]{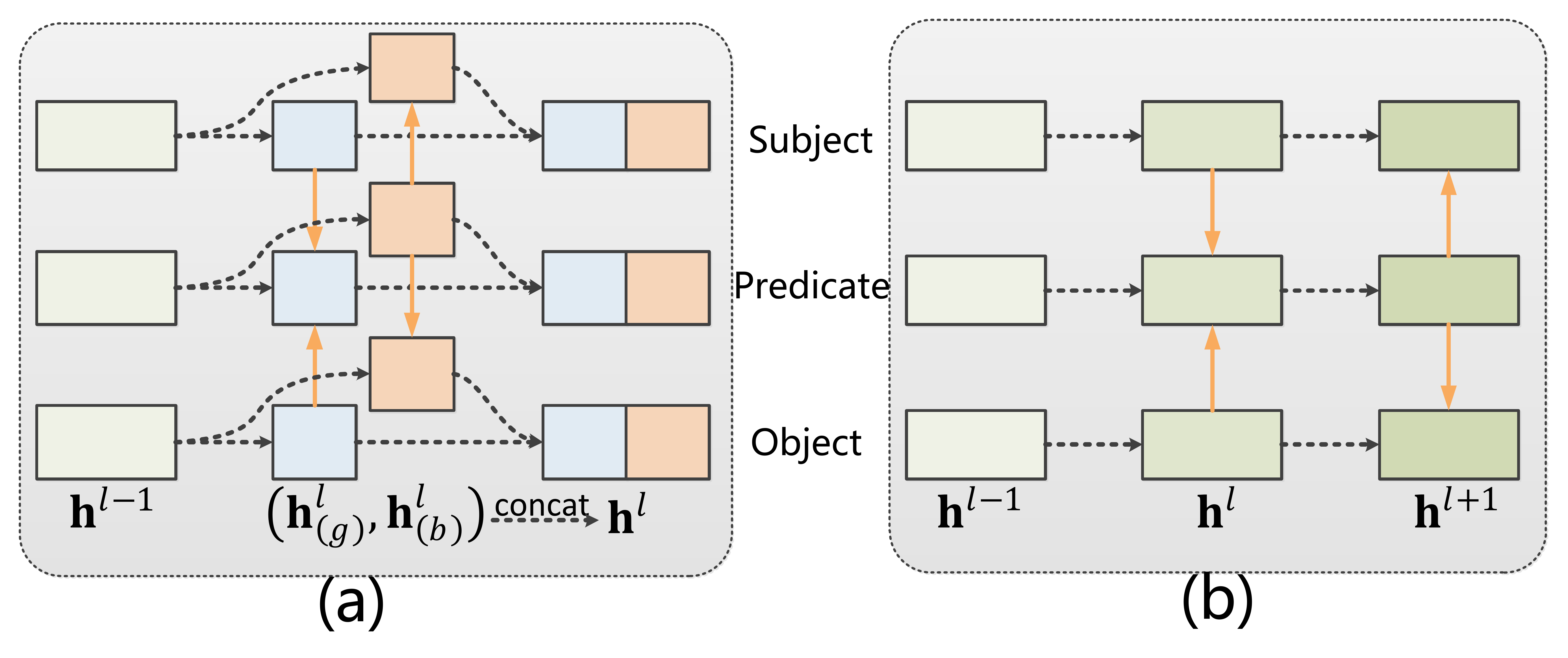}
		\end{center}
		\caption{Two types of PMPS. (a) \textbf{Parallel}; (b) \textbf{Sequential}. Sequential implementation needs two layers while the parallel implementation only needs one. $\mathbf{h}^l_{(g)} $ denotes features for gather flow, and $\mathbf{h}^l_{(b)} $ for broadcast.}
		\label{fig:mps}
	\end{figure}

	Subject-predicate-object triplet can be viewed as a simple graphical model. Predicate captures the general information about the phrase while subject and object focus on the details. To reflect the importance of the predicate within the triplet, we place the predicate at the dominant position and specifically design the gather-and-broadcast message passing flow. In the message passing flow, the predicate first gathers the messages from the subject and object  as follows:
	\begin{equation}
		\begin{split}
		\mathbf{h}_p^l = f\left(\mathbf{W}_p^l \otimes \mathbf{h}_p^{l-1} + \mathbf{W}_{p \leftarrow s}^l \otimes\mathbf{h}_s^l + \mathbf{W}_{p \leftarrow o}^l \otimes\mathbf{h}_o^l +\mathbf{b}_p^{l}\right). \label{eq:gather} \\
		\text{--- gather flow}
		\end{split}
	\end{equation}
	where $\mathbf{W}_{p \leftarrow s}^l$ and $\mathbf{W}_{p \leftarrow o}^l$  respectively denote the parameters for passing message from the subject and the object to the predicate.
	  At the next layer, the predicate broadcasts message to the subject and the object as follows:
		\begin{equation}
		\begin{split}
		\mathbf{h}_s^{l+1} = f\left(\mathbf{W}_s^{l+1} \otimes \mathbf{h}_s^{l} + \mathbf{W}_{s \leftarrow p}^{l+1} \otimes\mathbf{h}_s^{l+1} +\mathbf{b}_p^{l}\right), \\
		\mathbf{h}_o^{l+1} = f\left(\mathbf{W}_o^{l+1} \otimes \mathbf{h}_o^{l} + \mathbf{W}_{o \leftarrow p}^{l+1} \otimes\mathbf{h}_o^{l+1} +\mathbf{b}_p^{l}\right), \label{eq:broadcast} \\
		\text{--- broadcast flow}
		\end{split}
	\end{equation}
	where $\mathbf{W}_{* \leftarrow p}^{l+1}$ denotes the parameter used for passing message from the predicate features.
	
	In this PMPS, the gather flow collects the information from subject and object to refine the visual features of the predicate. Then, in the broadcast flow, the global visual information of interaction is broadcast back to the subject and object as context.
	
	The gather-and-broadcast flow has sequential and parallel implementations, as shown in Figure \ref{fig:mps}. The implementation in (\ref{eq:gather}) and (\ref{eq:broadcast}) is the sequential one, in which gathering is followed by broadcasting.
	In the parallel implementation, the original branch is divided into two parallel sub-branches, one for the gathering and the other for the broadcasting. The two sub-branches are concatenated after message passing.
	The sequential implementation is done on two adjacent layers while the parallel layer is done within the single layer. 
	
	%%\subsubsection{Apply MPS into Conv Layers}
	Since different branches have different semantic meanings in the subject-predicate-object structure, their learned features are different. We can also apply message passing for the convolutional layers before ROI pooling for mutual improvement. We empirically find that the sequential implementation is better for the fc layers and the parallel implementation is better for convolutional layers. %The comparison experiment results are shown in supplementary materials.  
	
	\begin{figure}[t]
		\begin{center}
			\includegraphics[width=1.0\linewidth]{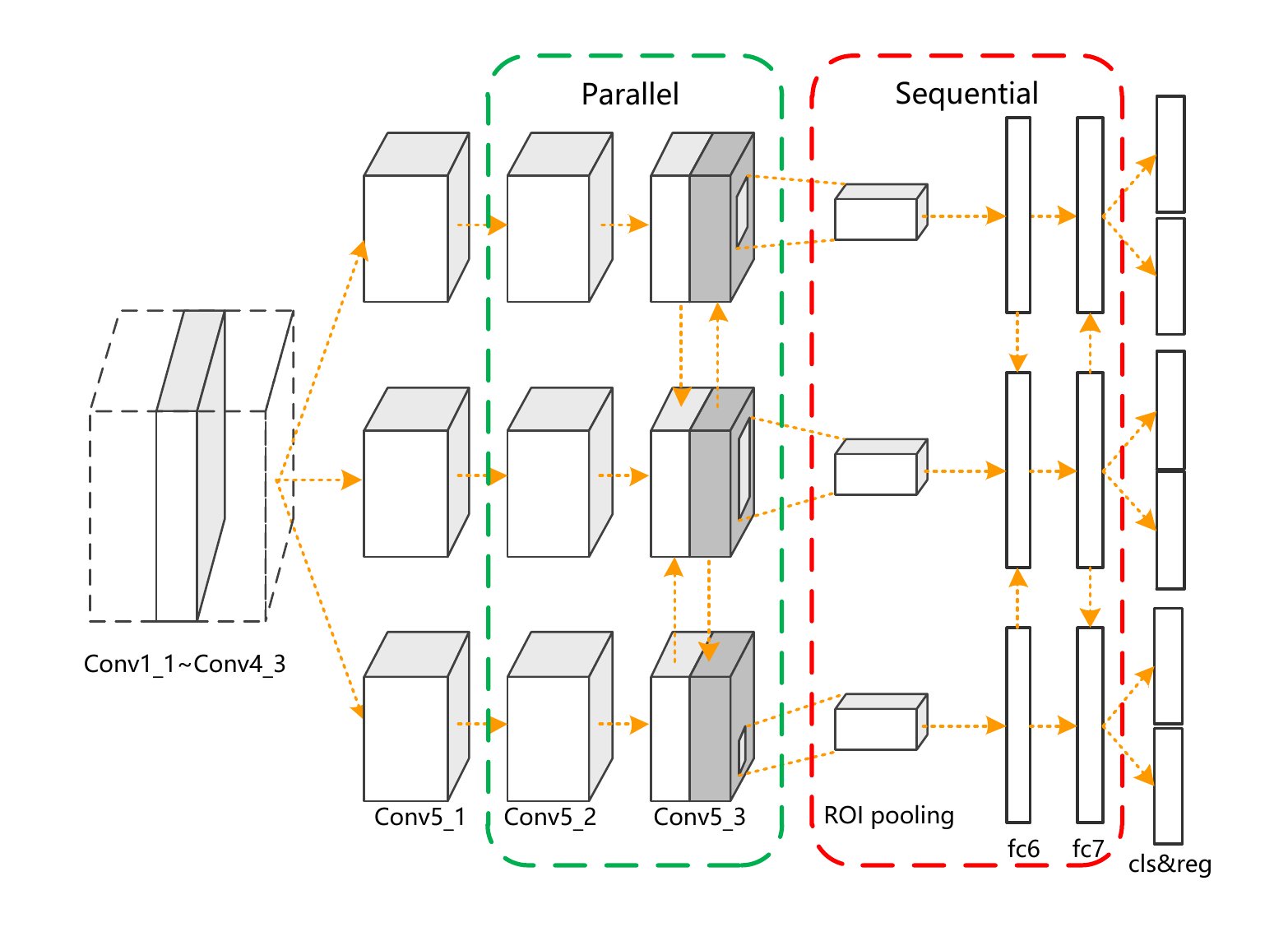}
		\end{center}
		\caption{Our final proposed model with PMPS on VGG-Net. Parallel implementation is placed at conv5\_3 layers while sequential one is placed at fc6 and fc7 layers.}
		\label{fig:final_model}
		\vspace{-10pt}
	\end{figure}

\section{Training the Model}

The triplet proposal and visual relationship detection are accomplished in a single network. At the training stage, we train it in multiple steps.

\subsection{Training  procedure}
For triplet proposal, we directly use the RPN proposed by He \etal. in ~\cite{faster_rcnn}. The model is initialized on ImageNet~\cite{imagenet} pretrained VGG-16 model, and then trained using subject and object instances.

For detection, the training process has two stages.

At the first stage, we remove the message passing structure PMPS and treat three branches as three separate detectors. We initialize each branch with ImageNet pretrained VGG-16 model and train the three branches at the same time. During training, the ROI in the triplet is marked as foreground when its overlap with corresponding ground truth is higher than a specific threshold (we use 0.5 for all three branches in the experiment), regardless of the other branches.

At the second stage, PMPS is enabled, and the model trained in stage one is used for initialization. In addition, to enforce the model considers the phrase as a whole, the subject/object/predicate ROI is regarded as foreground only if the overlaps of the subject, the object and the phrase proposals with their ground truth boxes are \emph{all} higher than the threshold. That is to say, even if the triplet ROIs correctly localize the phrase and subject, wrong localization of the object will force the subject and the predicate to be the background. Under this training strategy, the model is forced to consider the interdependency of the entire phrase. Therefore, this training strategy provides a top-down guidance, where the top message from the entire phrase is used for guiding the learning of three branches. 

To make the Conv1-Conv4 shared for triplet proposal and relationship detection, we take the four-step training like Faster R-CNN~\cite{faster_rcnn}. 

\subsection{Training  Loss}
Each branch has two outputs. For the subject branch, the output is the probability over the $ N+1 $ categories ($ N $ for targets and 1 for the background), $\mathbf{p}_s= \left( p_{s,0}, p_{s, 1}, ..., p_{s,N}\right) $,   and the corresponding box regression offsets, $\mathbf{t}_s = \left(t_{s,1}, t_{s,2}, ..., t_{s,N}\right) $. $t_{s,n} =\{t_{s,x}, t_{s,y}, t_{s,w}, t_{s,h}\} $, which denotes the scale-invariant translation and log-scale height/width shift on an given ROI~\cite{rcnn}.
	Similarly, we denote the probabilities and offsets for the predicate and the object as $\mathbf{p}_p$, $\mathbf{p}_o$, $\mathbf{t}_p$, $\mathbf{t}_o$.  We group the classification probabilities as 	$\mathbf{p} = \{\mathbf{p}_s, \mathbf{p}_p, \mathbf{p}_o\}$ and group the box regression offsets as $\mathbf{t} = \{\mathbf{t}_s, \mathbf{t}_p, \mathbf{t}_o\}$.
	The model employs multi-task loss for relationship detection\cite{fast_rcnn}:
\begin{equation}
	\begin{split}
	&L\left(\mathbf{p}, \mathbf{u}, \mathbf{t}^{\mathbf{u}}, \mathbf{v}\right) \\
	&= \sum_{\alpha \in \{s, p, o\}} L_{cls}(\mathbf{p}_\alpha, u_\alpha) + \lambda \left[ u_\alpha \ge 1\right]L_{reg}\left(\mathbf{t}^{u}_\alpha, \mathbf{v}_\alpha\right) ,
	\end{split}
\end{equation}
	where $ L_{cls} $ is the loss for classification, implemented by the soft-max loss, and $ L_{reg} $ is the bounding box regression loss, implemented using Smooth $ L_1 $ loss~\cite{fast_rcnn}. $ \mathbf{u}=\{u_s, u_p, u_o) $ denotes the targets for the subject, predicate and object. $ \mathbf{v}=\{\mathbf{v}_s, \mathbf{v}_p, \mathbf{v}_o\} $ denotes the ground truth regression values for the three proposals. The indicator function $ \left[u\ge 1\right] $ takes the value 1 for $ u\ge 1 $ and 0 otherwise, which means $ L_{reg} $ is ignored for the background.  The hyper-parameter $ \lambda $ balances the weights of the two losses.  $ \lambda=1$  in the experiment.

	\section{Experiment and Analysis}
	We use the pretrained VGG-16~\cite{VGG} provided by Caffe~\cite{caffe} to initialize our model. The newly introduced layers are randomly initialized. We set the base learning rate as 0.001 and fix the parameters from $ Conv1\_1$ to $Conv2\_2$. When finetuning the model on Visual Genome pretrained model, more parameters are fixed, from $Conv1\_1$ to $Conv4\_1$. For all experiments, we select top-200 boxes from region proposal and group them into triplets. 0.25 is used as the triplet NMS threshold.

	On single Titan X GPU, our model can finish the inference in about 0.6s per image. In comparison, the R-CNN-based state-of-art method~\cite{visual_relationship} requires more than 10s per image.

	\emph{Evaluation Criteria.} We adopt the Recall~\cite{alexe2012measuring} as the evaluation metric as in \cite{visual_relationship}. Top-N recall is denoted as Rec@N. For visual phrase detection, we aim to recognize  $ \langle subject$  - $predicate$ - $ object\rangle $ and  localize \emph{one} bounding box for the entire phrase. For relationship detection, we should recognize  $ \langle subject$  - $predicate$ - $ object\rangle $ and localize both subject and object. The proposal having at least 0.5 IOU with the ground-truth is regarded as correct localization.

	\subsection{Experiment on Visual Relationship dataset}

    Visual Relationship is proposed by Lu, \etal as a benchmark dataset for visual relationship detection. We will use the dataset to evaluate our proposed model and do the component analysis. 

	\subsubsection{Comparison to existing approaches} 
	
	We compare our model with the existing models~\cite{visual_phrase, visual_relationship}. 
	
	\begin{itemize}\setlength{\itemsep}{-.5pt }
		\vspace{-5pt}
		
		\item \textbf{Visual Phrases. } 6,672 deformable parts models are trained for every relationship category in training set of Visual Relationship dataset.
		
		\item \textbf{Language Prior. } Lu, \etal first use R-CNN~\cite{rcnn} to detect objects. A language model based on word vectors of the object categories and a visual model based on the CNN feature of the object pair are then trained to recognize the interactions.

		\item \textbf{Ours-ViP.} Full model of our proposed ViP-CNN as figure~\ref{fig:final_model} with weight sharing.
		\vspace{-5pt}
	\end{itemize}

	\begin{table}[h]
		\renewcommand{\arraystretch}{1.1}
		\setlength{\tabcolsep}{4pt}
		\small
		\begin{center}
			\begin{tabularx}{\linewidth}{l@{}cccc}
				\hline
				\multirow{2}{*}{Model} & \multicolumn{2}{c}{Phrase Det.} &  \multicolumn{2}{c}{Relationship Det.} \\
				& Rec@50& Rec@100 & Rec@50 & Rec@100 \\
				\hline
				Visual Phrases~\cite{visual_phrase} & 0.04 & 0.07 & - & - \\
				Language Prior~\cite{visual_relationship} & 16.17 & 17.03 & 13.86 & 14.70 \\
				\hline
				Ours-ViP & \textbf{22.78} & \textbf{27.91} & \textbf{17.32} & \textbf{20.01}\\
				\hline
			\end{tabularx}
		\end{center}
		\caption{Evaluation of different methods at Visual Relationship~\cite{visual_relationship} on visual phase detection (Phrase Det.) and visual relationship detection (Relationship Det.) measured by Top-50 recall (Rec@50) and Top-100 recall (Rec@100).}
		\label{tab:model}
	\end{table}

	Visual Phrases~\cite{visual_phrase} performs poorly due to the lack of training instances. Compared with the state-of-art method that uses Language Prior~\cite{visual_relationship}, our model increases Recall@100 by 9.04\% and 4.38\% on visual phrase detection and visual relationship detection task respectively. Since both models are based on VGG-Net~\cite{VGG}, the gain is mainly from the better use of the interdependency within the phrase.

	\subsubsection{Component Analysis}

	\begin{table}[t]
		\renewcommand{\arraystretch}{1.1}
		\setlength{\tabcolsep}{4pt}
		\small
		\begin{center}
			\begin{tabularx}{\linewidth}{l@{}cccc}
				\hline
				\multirow{2}{*}{Model} & \multicolumn{2}{c}{Phrase Det.} &  \multicolumn{2}{c}{Relationship Det.} \\
				& Rec@50& Rec@100 & Rec@50 & Rec@100 \\
				\hline
				Baseline & 13.69& 16.41 & 10.31 & 12.75\\
				Baseline-Concat & 15.71 & 18.90 & 11.98 & 14.33 \\
				RNN & 14.08 & 18.01 & 11.08 & 13.25 \\
				\hline
				ViP-No NMS & 10.68 & 16.28 & 9.01 & 11.87 \\
				ViP-Rand. Select & 17.71 & 23.66 & 13.96 & 17.04\\
				ViP & 21.24 & 26.07 & 16.57 & 19.08\\
				ViP-Post NMS & 22.31 & 27.24 & 16.95 & 19.81\\
				ViP-Param Sharing & \textbf{22.78} & \textbf{27.91} & \textbf{17.32} & \textbf{20.01}\\
				\hline
			\end{tabularx}
		\end{center}
		\caption{Evaluation of different methods at Visual Relationship~\cite{visual_relationship} on visual phase detection (Phrase Det.) and visual relationship detection (Relationship Det.) tasks measured by Top-50 recall (Rec@50) and Top-100 recall (Rec@100).}
		\label{tab:component}
		\vspace{-8pt}
	\end{table}
	
	There are many components that influence the performance of the proposed approach. Table \ref{tab:component} shows our investigation on the performance of different settings on the visual relationship dataset~\cite{visual_relationship}.

	\emph{Phrase-guided Message Passing Structure.}
	\textbf{Baseline} model disables PMPS and the three models make prediction separately. Compared to our full ViP-CNN (\textbf{ViP} in Figure~\ref{tab:component}), there is 10.66\% and 6.33\% Rec@100 drop on visual phrase detection and visual relationship detection respectively. \textbf{Baseline-Concat} directly concatenates the three branches to make them fully connected at feature level. Instead of manually designing the information flow, it allows the model to learn that by itself. However, it underperforms ViP by 7.17\% and 4.71\%  Rec@100 on the two tasks, which shows that the use of domain knowledge can facilitate the model training. \textbf{RNN} model utilizes the feature-level interdependency with internal memory structure. But its weight sharing scheme and fixed flow path deteriorate the final results. In comparison, our ViP-CNN outperforms RNN by 8.06\% and 5.83\% Rec@100 on the two tasks respectively. Experimental results show that PMPS can help our proposed model better use the inter-connection among the phrase components for more accurate inference. 
	
	\emph{Triplet NMS.}
	In our final model, triplet NMS is used before detection. We also investigate removing this NMS procedure and then randomly selecting 2000 triplets from 62,500 potential triplets to feed into detection network (\textbf{ViP-Rand. Select}).  It can be seen that random sampling of triplets leads to 2.41\% Rec@100 drop on visual phrase detection task when compared with our full ViP-CNN model which uses the proposed triplet NMS.  On the other hand, removing NMS from the pipeline performs worse than random selection, with 7.38\% Rec@100 reduction for visual phrase detection  (\textbf{ViP-No NMS}). 
	Furthermore, when NMS is moved placed posterior to the detection, which is denoted by \textbf{ViP-Post NMS}, Rec@100 can increase by 1.14\% and 0.74\% on phrase detection and relationship detection. However, the execution time becomes more than 20 times as much as before. Thus, we adopt pre-detection NMS like Figure~\ref{fig:pipeline} to balance the performance and speed. 
	
	\emph{Weight Sharing in ViP-CNN. } Object and subject branch share most of the properties. They detect the same set of objects, and they are of the similar status when we view the entire relationship as a graphical model. Therefore we can share the parameters for the subject and object branches except for their message passing parameters, denoted by \textbf{Ours-ViP+Param Sharing}. It can help the model to learn more general features and reduce parameter size. 
	The experiment result shows that the parameter sharing scheme helps to improve Rec@100 of ViP-CNN by 1.84\% and 0.93\% for visual phrase detection and visual relationship respectively.

	\subsection{Experiments on Visual Genome}

	Newly-introduced dataset, Visual Genome~\cite{visual_genome}, has several kinds of annotations, one of which is visual relationships. We denote the Visual Genome Relationship dataset as VGR. The dataset contains 108,077 images and 1,531,448 relationships. 
	
	However, we find that the annotations of VGR contain some misspellings and noisy characters (\eg comma). And verbs and nouns are also in different forms. Therefore, by cleansing the Visual Genome Relationship dataset~\cite{visual_genome}, we build up a new relationship dataset, denoted as VGR-Dense. %The detailed criteria will be shown in Supplementary Material. 
	
	Due to the long-tail distribution of the categories in VGR-Dense, we further filter out the infrequent object and predicate categories by setting 200 as the frequency threshold for object categories and 400 for the predicate categories. Then we get another visual relationship dataset with more frequent categories, which is denoted as VGR-Frequent.

	We will test our proposed model on these two datasets to further evaluate our proposed model. In addition, we will investigate different ways of pretraining methods on the two datasets for visual phrase/relationship detection task.

	We randomly divide the dataset into three subsets, 70\% as the training subset, 10\% as the validation subset and 20\% as the testing subset. Model training is done on the training subset. Additionally, the images overlapping with Visual Relationship~\cite{visual_relationship} are removed from the training set.

	\begin{table}[t]
		\renewcommand{\arraystretch}{1.1}
		\small
		\begin{center}
			\begin{tabular}{lcccc}
				\hline
				Dataset & \#Images &\#Rel & \#Obj & \#Pred \\
				\hline
				VR~\cite{visual_relationship} & 5,000 & 37,993 & 100 & 70 \\
				VGR~\cite{visual_genome} & 108,077 & 1,531,448 & - & - \\
				ours-VGR-Dense  &102,955& 1,066,628 & 3407 & 142\\
				ours-VGR-Frequent  &101,649& 900,739 & 507 & 92 \\
				\hline
			\end{tabular}
		\end{center}
		\caption{Statistics on the Visual Relationship~\cite{visual_relationship} (VR), Visual Genome Relationship dataset~\cite{visual_genome} (VGR) and two cleansed dataset based on Visual Genome. The number of images (\#Images), relationship triplets (\#Rel), object categories (\#Obj) and predicate categories (\#Pred) are shown.}
		\label{tab:dataset}
		\vspace{-15pt}
	\end{table}

	\subsubsection{Model Evaluation on Visual Genome Relationship dataset}

	Since there are no existing works reported on Visual Genome dataset, we use our proposed model without PMPS as baseline model, denoted as \textbf{Baseline}, and compare our ViP-CNN with it to investigate the effect of our proposed PMPS and weight sharing strategy (\textbf{ViP} denotes our proposed ViP-CNN without parameter sharing in figure~\ref{fig:final_model}, \textbf{ViP-P.S.} denotes the ViP-CNN with parameter sharing). 
	
		\begin{table}[h]
		\renewcommand{\arraystretch}{1.1}
		\setlength{\tabcolsep}{4pt}
		\small
		\begin{center}
			\begin{tabularx}{\linewidth}{l@{}lcccc}
				\hline
				\multirow{2}{*}{Dataset} & \multirow{2}{*}{Model} & \multicolumn{2}{c}{Phrase Det.} &  \multicolumn{2}{c}{Relationship Det.} \\
				& & Rec@50& Rec@100 & Rec@50 & Rec@100 \\
				\hline
				\multirow{3}{1.4cm}{VGR Dense}&Basline & 7.32 & 9.98 & 4.01 & 7.89 \\
				&ViP & 13.71 & 16.75 & 8.12 & 11.81   \\
				&ViP-P.S. & \textbf{14.17} & \textbf{17.29} & \textbf{8.74} & \textbf{12.17}\\
				\hline\hline
				\multirow{3}{1.4cm}{VGR Frequent}& Basline & 8.17 & 13.89 & 5.71 & 8.94 \\
				& ViP & 16.03 & 20.85 & 9.98 & 12.90\\
				& ViP-P.S. & \textbf{16.58} & \textbf{21.54} & \textbf{10.67} & \textbf{13.81}\\
				\hline
			\end{tabularx}
		\end{center}
		
		\caption{Evaluation of different methods on Visual Genome~\cite{visual_genome} for visual phase detection (Phrase Det.) and visual relationship detection (Relationship Det.) tasks measured by Top-50 recall (Rec@50) and Top-100 recall (Rec@100).}
		\label{tab:VGR}
		\vspace{-6pt}
	\end{table}
	
	The experiments on the two Visual-Genome-based relationship datasets both proves the effect of our proposed PMPS. For VGR-Dense, 6.77\% and 3.92\% Rec@100 increase on visual phrase detection and visual relationship detection come from the introduction of PMPS, while for VGR-Frequent, the increases are 6.96\% and 3.96\% correspondingly.
	
	Furthermore, we evaluate our proposed parameter sharing strategy. On VGR-Dense, it brings 0.54\% and 0.36\% Rec@100 gains on the two tasks respectively. On VGR-Frequent, the gains are 0.68\% and 0.91\%. The results show that parameter sharing still works on the large-scale relationship dataset, although the gain is not as much as on the small set like Visual Relationship~\cite{visual_relationship}.

	\subsubsection{Investigation on pretraining settings}\label{sec:vg}
	
	We further investigate how the training data and training targets influence the pretraining on our two datasets.  The results tested on the visual relationship dataset are shown in Table \ref{tab:pretrain}.
	
	For pretraining dataset, the \textbf{baseline} is pretrained on the ImageNet Classification data, the others are pretrained on our VGR-dense and VGR-frequent. The pretrained models are used for initialization and then finetuned on the Visual Relationship dataset~\cite{visual_relationship}. During finetuning, the parameters from $ Conv1_1 $ to $ Conv4_1 $ are all fixed, because it is found not influencing results in \cite{fast_rcnn}.
	
	The categories in the two datasets follow long-tail distribution. Especially for the VGR-Dense dataset, most of the classes have few instances for training. Therefore, we can use some structured targets to make full use of the instances of rare categories for pretraining. Because of the amazing properties of word vector~\cite{word2vec}, we can convert the label to their word vectors as the training targets to make full use of the rare categories. For word vector, we employ the Smooth $ L_1 $ loss function~\cite{fast_rcnn}, which is denoted by \emph{vec} in Table \ref{tab:pretrain}. In addition, we also use the original multi-class targets as comparison, which is denoted by \emph{class}.
	
	\begin{table}[h]
		\vspace{-7pt}
		\renewcommand{\arraystretch}{1.1}
		\setlength{\tabcolsep}{4pt}
		\small
		\begin{center}
			\begin{tabularx}{\linewidth}{p{1.4cm}@{}ccccc}
				\hline
				\multirow{2}{1.4cm}{Pretrain Dataset} & \multirow{2}{*}{Target} & \multicolumn{2}{c}{Phrase Det.} &  \multicolumn{2}{c}{Relationship Det.} \\
				& & Rec@50& Rec@100 & Rec@50 & Rec@100 \\
				\hline
				baseline & - & 22.78 & 27.91 & 17.32 & 20.01\\
				\hline
				\multirow{2}{1.4cm}{Dense}& vec & 23.34& 29.56 & 18.42 & 21.37 \\
				& class & 23.98 & 30.01 & 19.01 & 21.96 \\
				\hline
				\multirow{2}{1.4cm}{Frequent}&vec & 23.29 & 29.61 & 18.39 & 21.32 \\
				& class & \textbf{24.21} & \textbf{30.51} & \textbf{19.44} & \textbf{22.28} \\
				\hline
			\end{tabularx}
		\end{center}
		\caption{Comparison of different pretraining methods. \emph{Baseline} is pretrained on ImageNet.  All models are finetuned and tested on the dataset in \cite{visual_relationship}. Target denotes the training targets for classification, class label (\emph{class}) or word vector (\emph{vec}).}
		\label{tab:pretrain}
		\vspace{-7pt}
	\end{table}

	From the experimental result, we can see that pretraining using our cleansed datasets,  VGR-Dense and  VGR-Frequent, performs better than the pretraining on ImageNet.  When the class label is used as the target, pretraining using the VGR-Frequent dataset has 22.28\% Rec@100 on visual relationship detection, is the best choice, with 2.27\% gain when compared with the baseline. Pretraining using the VGR-frequent dataset slightly outperforms pretraining using the VGR-Dense dataset, with 0.32\% Rec@100 improvement. Category labels perform better than word vector labels for pretraining, with 0.96\% Rec@100 improvement on visual relationship detection when using the VGR-frequent dataset. Besides, with word vector supervision, the two datasets have similar performance.
	
	Based on the results, too many low-frequency categories will deteriorate the pretraining gain when utilizing class label as training targets. With implicitly embedded structure, word vector is expected to be a  solution to the problem. However, the result reveals that under current experiment settings, it still cannot surpass the widely used multi-class label targets. But the word vector target is still a possible way to utilize the large quantities instances of rare classes, which is worthwhile for other potential applications.
	
	\begin{table}[t]
		\vspace{-5pt}
		\renewcommand{\arraystretch}{1.1}
		\setlength{\tabcolsep}{4pt}
		\small
		\begin{center}
			\begin{tabularx}{\linewidth}{lccc}
				\hline
				model & 	Faster R-CNN~[9] & Ours-baseline & ViP-CNN \\
				\hline
				mean AP (\%) & 14.35 & 14.28 & \textbf{20.56}\\
				\hline
			\end{tabularx}
		\end{center}
		\vspace{-5pt}
		\caption{\textbf{Object detection} result on the dataset in \cite{visual_relationship}. Ours-baseline has removed message passing.
		}
		\label{tab:object_detection}
		\vspace{-15pt}
	\end{table}
	
	\subsection{Further investigation on Object Detection}
	To separately investigate the improvement on recognizing objects, we further compare ViP-CNN with Faster R-CNN~\cite{faster_rcnn} and the baseline model without PMPS on the intersection of the subject and object sets. ViP-CNN only uses the \emph{subject} detection results without any specific tuning.     
	Our proposed model outperforms Faster R-CNN and our baseline model by 6.21\% and 6.28\% respectively measured by mAP (Table~\ref{tab:object_detection}). By comparing the three results, we can see that the gain mainly comes from the extra context information introduced by PMPS. Therefore, how to leverage the context information with explicit semantic meaning based on the visual relationship dataset could be a possible direction in the future work. 

	\section{Conclusion}

	In this paper, our proposed Visual Phrase Guided Convolutional Neural Network is proved to be effective for visual relationship detection. A triplet NMS procedure is proposed to remove redundant detection results for more efficient inference. In ViP-CNN, a message passing structure called PMPS is proposed to model the visual interdependency among relationship components, which setting up a horizontal information flow among different deep models at the same layer. PMPS helps deep models for s/o/p improve one another through mining their connections. Evaluated on the Visual Relationship dataset, our model outperforms the state-of-art model in both speed and accuracy.
	Experimental results of the pretrained model on Visual Genome Relationship dataset are also presented. It performs better than the ImageNet pretrained model on the visual phrase/relationship detection task. In comparison, the word vector training target is not comparable with the label target under the present settings. Future work can be done in this direction for mining the structural information among the labels.
	
	\section*{Acknowledgement}
	This work is supported by Hong Kong Ph.D Fellowship scheme, Sense-Time Group Limited, the General Research Fund sponsored by the Research Grants Council of Hong Kong (Project Nos. CUHK14213616, CUHK14206114, CUHK14205615, CUHK419 412, CUHK14203015, and CUHK14207814), the Hong Kong Innovation and Technology Support Programme (No.ITS/121/15FX), National Natural Science Foundation of China (Nos. 61371192, 61301269), and PhD programs foundation of China (No. 20130185120039).
	We also thank Xiao Tong, Kang Kang and Xiao Chu for helpful discussions along the way.

	{\small
		\bibliographystyle{ieee}
		\bibliography{egbib}
	}

\end{document}